\documentclass[letterpaper, 10 pt, conference]{ieeeconf}  

\usepackage{graphicx}
\usepackage{xcolor}
\usepackage{colortbl}
\usepackage{epstopdf}
\usepackage{subfigure}
\usepackage{amsmath}
\usepackage{hyperref}
\usepackage{float}
\usepackage[linesnumbered,ruled,vlined]{algorithm2e}
\usepackage{cite}
\usepackage{amssymb}

\hypersetup{hidelinks,
	colorlinks=true,
	allcolors=black,
	pdfstartview=Fit,
	breaklinks=true}

\IEEEoverridecommandlockouts                              
\title{\LARGE \bf
LU2Net: A Lightweight Network for Real-time \\ Underwater Image Enhancement  
}

\author{Haodong Yang, Jisheng Xu, Zhiliang Lin and Jianping He
\thanks{Haodong Yang is with the  Department  of  Computer Science,  Shanghai  Jiao  Tong University. Jisheng Xu, Zhiliang Lin and Jianping He  are  with  the  Department  of  Automation,  Shanghai  Jiao  Tong University,  Key  Laboratory  of  System  Control  and  Information  Processing,  Ministry  of  Education  of  China,   and Shanghai Engineering Research Center of Intelligent Control and Management, Shanghai  200240,  China. 
        Zhiliang Lin is with the School of Ocean and Civil Engineering, Shanghai Jiao Tong University, State Key Laboratory of Ocean Engineering, Shanghai 200240, China. Emails: 
        {
        \{yanghaodong, Jimmy\_xu, linzhiliang, jphe\}@sjtu.edu.cn.
        }
}        
}

\begin{document}

\maketitle
\thispagestyle{empty}
\pagestyle{empty}
\makeatletter
\newcommand{\rmnum}[1]{\romannumeral #1}
\newcommand{\Rmnum}[1]{\expandafter\@slowromancap\romannumeral #1@}
\makeatother

\begin{abstract}
Computer vision techniques have empowered underwater robots to effectively undertake a multitude of tasks, including object tracking and path planning. However, underwater optical factors like light refraction and absorption present challenges to underwater vision, which cause degradation of underwater images. A variety of underwater image enhancement methods have been proposed to improve the effectiveness of underwater vision perception. Nevertheless, for real-time vision tasks on underwater robots, it is necessary to overcome the challenges associated with algorithmic efficiency and real-time capabilities. 
In this paper, we introduce Lightweight Underwater Unet (LU2Net), a novel U-shape network designed specifically for real-time enhancement of underwater images. The proposed model incorporates axial depthwise convolution and the channel attention module, enabling it to significantly reduce computational demands and model parameters, thereby improving processing speed. The extensive experiments conducted on the dataset and real-world underwater robots demonstrate the exceptional performance and speed of proposed model. It is capable of providing well-enhanced underwater images at a speed 8 times faster than the current state-of-the-art underwater image enhancement method. Moreover, LU2Net is able to handle real-time underwater video enhancement.

\end{abstract}

\section{INTRODUCTION}
With the remarkable progress of computer vision in recent years, underwater vision is emerging as a prevalent and vital 
means of gathering information for underwater robots. A variety of vision models have been integrated into underwater robots
for marine tasks, like object detection \cite{ref35}, monocular depth estimation \cite{ref39} and visual odometry \cite{ref33}.
Real-time and high-quality underwater images provide essential information for robotic decision making and improve the safety
and stability of underwater robots.


Though computer vision technologies strengthen the perception abilities of underwater robots, low-quality
underwater images degrade the performance of marine robotic tasks. 
In contrast to the relatively stable onshore environment, the underwater environment is characterized by complexity and variability.
Greater absorption of red light and 
scattering of suspended particles in water lead to imbalanced color channels and inconsistent blurriness in images \cite{ref36}. 
 Additionally, light sources may be insufficient in deep water, limiting the visibility of target objects and surroundings. 
The changeable underwater environment hugely affects the quality of underwater images. Thus, the computer vision models designed for 
onshore clear images and videos may suffer from decrease in performance when applied in underwater robots for marine vision tasks.
        
        
    
 


To restore degraded underwater images, researchers have proposed a variety of underwater image enhancement (UIE) methods. Traditional UIE Methods that are based on visual and physical prior \cite{ref3,ref4,ref5,ref6,ref40,ref41,ref7,ref8}
are usually unable to adapt to various underwater environments and result in inappropriate enhancement. 
Deep neural networks for underwater image enhancement has been intensively investigated. UIEC\^{}2-Net \cite{ref42} integrates RGB and HSV color spaces in one single 
CNN, utilizing different image properties.  
PUGAN \cite{ref43} estimates physical parameters to guide image enhancement in neural networks.
Recently, \cite{ref13} introduced transformer into UIE. 
However, the above-mentioned learning-based methods often focus on visual effects rather than processing speed. For underwater robots performing practical marine tasks, it is necessary to provide fast image enhancement.

To address the aforementioned issue, in this paper, we propose \textbf{L}ightweight \textbf{U}nderwater \textbf{U}Net (LU2Net) 
for real-time underwater images enhancement. 
Axial depthwise convolution is embedded into LU2Net, which contributes to 
larger receptive fields. Thus more details are perceived with less convolution layers. Besides, channel 
attention module is integrated to adaptively adjust channel weights and mitigate inconsistent attenuation among channels. 
The experiments on the dataset and real-world robots demonstrate that our model outperforms state-of-the-art methods in both quantitative metrics and processing speed.
Our main contributions can be summarized as:
\begin{figure*}[ht]
    \centering
    \includegraphics[scale=0.49]{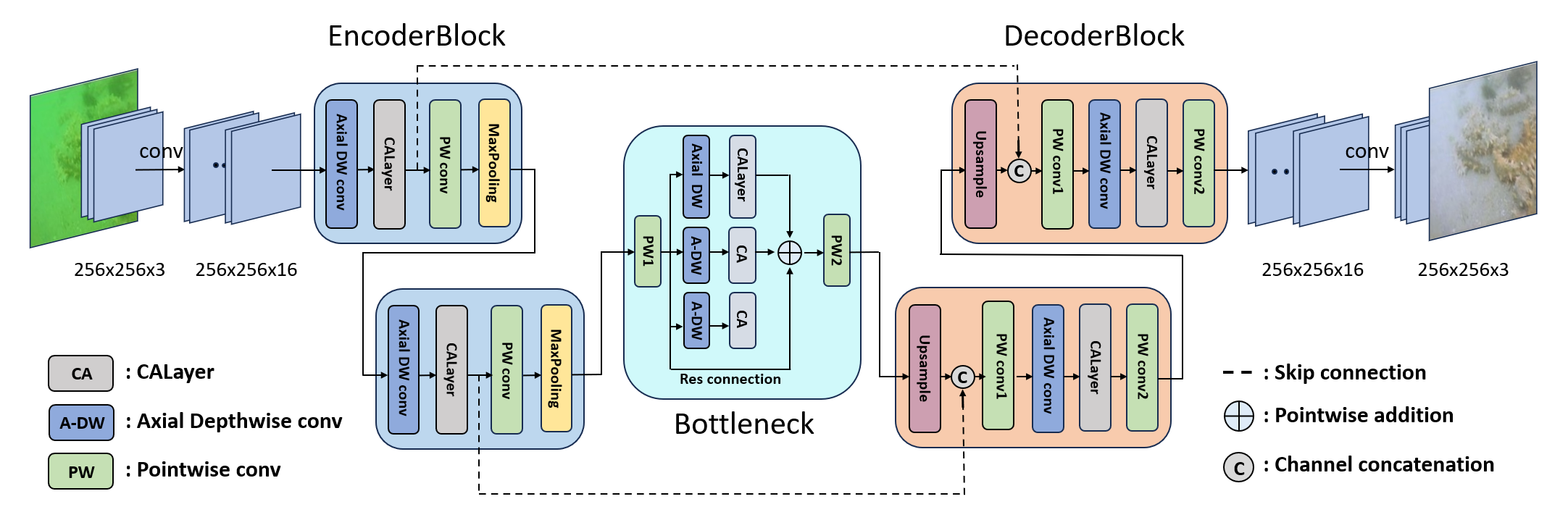}
    \vspace{-20pt}
    \caption{The illustration of LU2Net structure. Specially designed encoder and decoder blocks enable larger receptive fields and the adaptive adjustment of 
    channel weights. Skip connection ensures the full utilization of multi-stage information.}
    \label{fig-net_structure}
\end{figure*}
\begin{itemize}
    \item[$\bullet$] We propose a novel U-shape network, LU2Net, that integrates a lightweight convolution and adaptive channel weights for real-time underwater image enhancement. 
                     This lightweight U-shape model enables underwater robots to obtain high-quality and real-time images for marine vision tasks. 
    \item[$\bullet$] 
                     Our specially designed block structure is embedded with channel-wise attention and axial convolution.
                     This simple and fast block structure mitigates inconsistent attenuation in color channels and achieves large receptive fields, consuming a small amount of time and resources. 
    \item[$\bullet$] 
                     Our model demonstrates significant advantages through experiments on a dataset and practical testing on underwater robots.
                     Maintaining better performance, our model achieves
                     8 times faster speed than state-of-the-art model, showing LU2Net's ability on videos.
\end{itemize}


\section{Related Works}

\subsection{UIE Methods}
Various methods have been proposed to enhance recorded single underwater images, which can be divided into the following three categories.

i) \textbf{Visual prior methods:} These approaches focus on adjusting pixel values to improve image quality. Early methods include histogram equalization \cite{ref1}, which transforms the image histogram from a
narrow unimodal histogram to a balanced distribution histogram. Image fusion \cite{ref2,ref10} takes enhanced images by different algorithms as input and generate results by fusing them.  
Retinex-based methods \cite{ref6} obtain the actual appearance of the object by estimating and removing the illumination light in the environment. 
More recently, morphology-based techniques \cite{ref4,ref5} have been introduced to underwater image enhancement. Visual prior methods inherently improve contrast and saturation 
in underwater images. However, they may lead to over-enhancement when applied to real-world underwater images.

ii) \textbf{Physics-based methods:} These methods utilize prior knowledge of the physics of underwater imaging to guide image enhancement. For example, UDCP algorithm \cite{ref7} is based on the fact that most non-sky pixels in haze-free images possess at least one color channel with significantly low intensity. By estimating the weights of different channels through less attenuated blue and green channels, UDCP  demonstrates effectiveness in processing deep-sea images. Another approach,
RDCP \cite{ref8}, leverages the information of fastly decayed red light and considers both natural light and artificial light to achieve good enhancement results. However, as these methods focus on physical models and do not consider human visual perception, the visual quality of the enhanced images may be compromised.

iii) \textbf{Deep learning methods:} The emergence of deep learning has sparked interest in data-driven underwater image enhancement. CNN-based methods 
learn the mapping from original underwater images to high-quality images.
UWCNN \cite{ref12} introduced a simple CNN architecture with stacked convolution layers for UIE tasks. 
UIEC\^{}2-Net \cite{ref42} utilizes image properties in various color spaces. Besides, 
a variety of GAN-based models have been proposed, which usually take non-image information for enhancement.
WaterGAN \cite{ref9} estimates and utilizes depth information. 
PUGAN \cite{ref43} combines the physical model of underwater imaging and the generative adversarial network.  
Recently, transformer-based methods \cite{ref13} are also introduced into UIE.
Though pleasing visual effect is achieved, the importance of speed and net size is frequently ignored. Current deep learning methods often suffer from relatively slow processing speed and a reliance on high-end graphics processing hardware.
This hinders their applicability for real-time vision tasks on robotic platforms with limited computing capabilities.

\subsection{Light Weight U-Net}
U-Net  
employs a U-shape network with encoder-decoder architecture for biomedical image segmentation \cite{ref21}.
After the emergence of U-Net, the U-shape architecture has been widely adopted in various vision tasks. For instance, it has 
proven effective for image enhancement. Peng et al. \cite{ref13} introduced U-shape structure into UIE task. 

After the success of U-Net, researchers have explored methods to 
streamline U-shape networks, aiming to reduce computational complexity and improve efficiency. For example, Unext proposed in \cite{ref24} combines a multi-layer perceptron with U-Net, resulting in less convolution layers and network parameters. 
In \cite{ref25}, U-Lite  leverages depth separable convolution, significantly reducing the parameter count while maintaining high performance. Motivated from these works, in this paper, we utilize the idea from U-Lite for reduction of network size and complexity. 

\addtolength{\textheight}{-3cm}   

\section{METHODOLOGY}
In this section, we introduce the structure of LU2Net. As illustrated in Fig. \ref{fig-net_structure}, U-shape network structure is utilized for 
multi-scale information extraction. 
A lightweight convolution architecture, axial depthwise convolution \cite{ref25}, is 
leveraged to achieve larger receptive fields with less layers and parameters. Besides, the channel attention module proposed in \cite{ref34} 
is embedded in LU2Net, which adaptively adjust channel weights and effectively correct color distortion. 

With these modules, novel encoder and decoder block structures are designed for improved performance and 
less parameters and 
shortened processing time. The novel block structures allow for reduced size and reinforced speed of our model while 
maintaining significant enhancement performance, which is illustrated in Section \Rmnum4.

\subsection{U-Shape Network Architecture}
As is shown in Fig. \ref{fig-net_structure}, LU2Net proposed in this paper adopts a streamlined U-shape network design, achieving high performance without excessive block stacking. Our U-shape network architecture consists of an classic encoder-decoder structure. Stacked encoder and decoder blocks provide a holistic view of multi-scale features. 
The skip connections in the network facilitate the transfer of information from earlier encoder stages to later decoder stages, improving the performance by incorporating multi-stage information.


Different from traditional U-Net \cite{ref21}, whose blocks are constructed by stacked convolution layers, the proposed model introduces novel encoder and decoder block structures. The blocks incorporate axial depthwise convolution, enabling 
larger receptive fields. Then, less convolution layers are required in our blocks.
Channel attention module in blocks provide adaptive adjustment of channel weights, which provides a light and fast solution for underwater color distortion. 
Utilizing this block structures, LU2Net achieves state-of-the-art enhancement performance and outstanding processing speed
with a small number of blocks. 

\subsection{Axial Depthwise Convolution Module}
Axial depthwise convolution is integrated in our model for higher performance while reducing parameters and computational complexity. The axial depthwise convolution combines depthwise separable convolution \cite{ref44} and axial feature extraction, resulting in a novel convolution variant\cite{ref25}. By replacing common convolution layers with axial depthwise convolution, the amount of parameters in blocks largely decreases while the performance is improved.  

\begin{figure}[ht]
    \centering
    \includegraphics[scale=0.4]{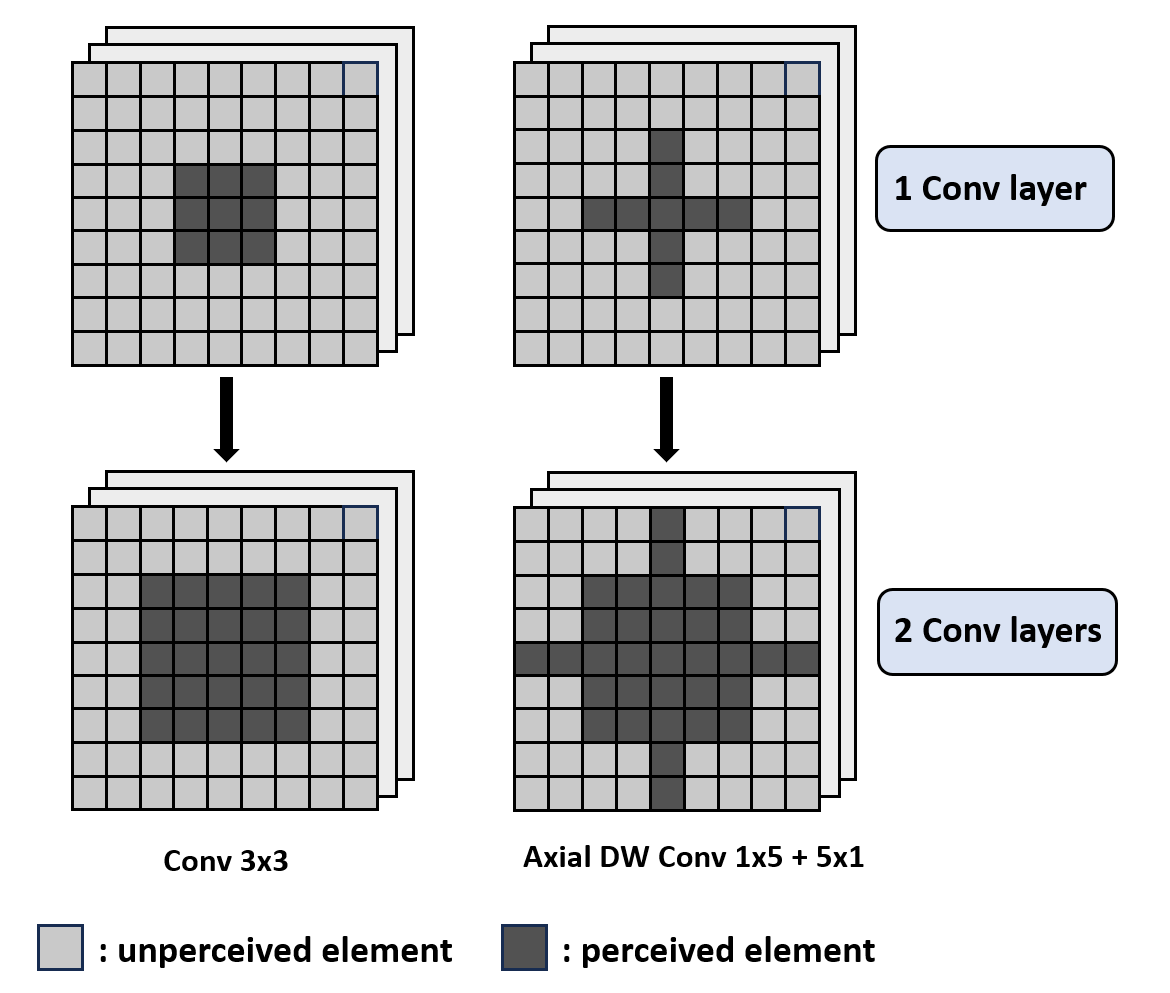}
    \caption{The comparsion of receptive field between axial depthwise convolution and traditional convolution. As the network get deeper with 
    more convolution layers, the improvement of receptive fields in axial depthwise convolution is strengthened. Thus more details are perceived 
    with the number of layers unchanged. }
    \label{fig-adw}
\end{figure}

By transforming traditional square convolution kernels into axial variants, the receptive fields of axial depthwise convolution expand while 
reducing the number of parameters. Fig. \ref{fig-adw} showcases the larger receptive fields of axial depthwise convolution compared to traditional convolution. This modification allows 
for leveraging more precise details from the input image, leading to superior performance with fewer parameters. 

\begin{figure}[ht]
        \centering
        \includegraphics[scale=0.5]{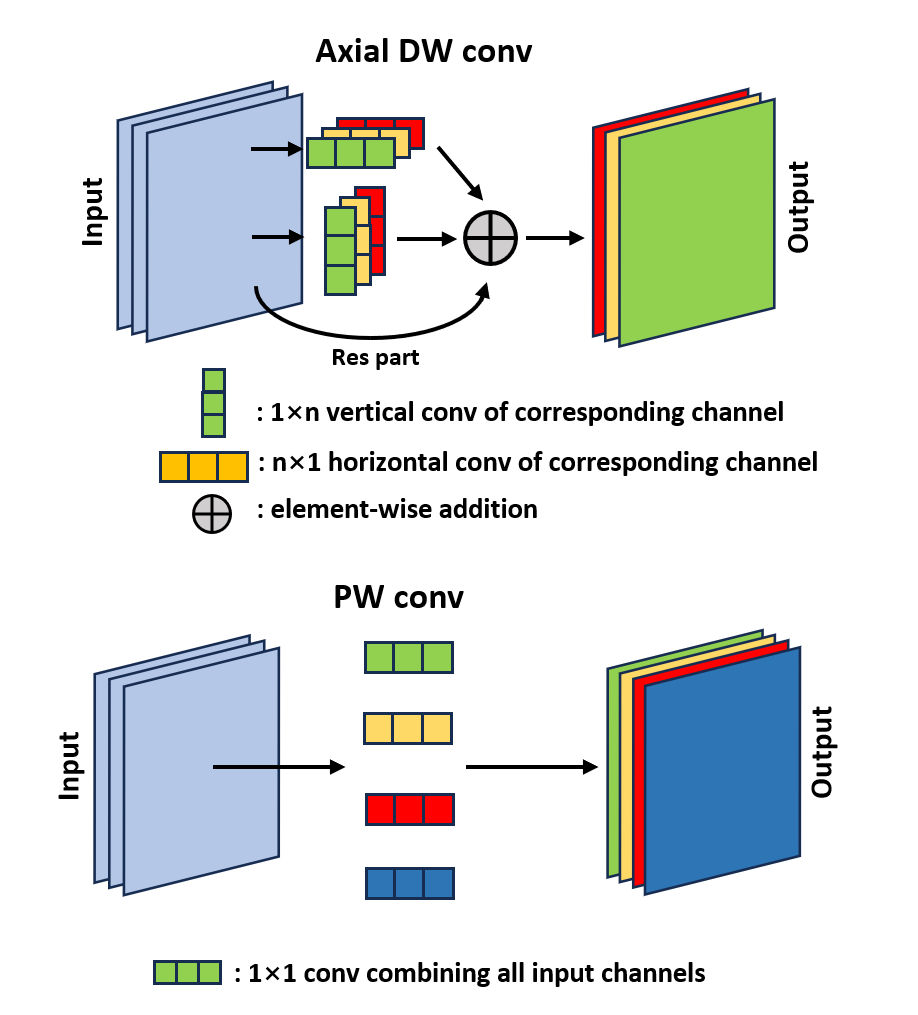}
        \caption{The structure of axial depthwise convolution and pointwise convolution. In axial depthwise convolution, each convolution kernel operates 
        on single input channel and the number of channel remains the same. While in pointwise convolution, each kernel combines all input channels and corresponds 
        to an output channel.}
        \label{fig-DW_PW}
\end{figure}

 Axial depthwise convolution consists of two stages, as is shown in Fig. \ref{fig-DW_PW}. Depthwise convolution keeps the depth of the output unchanged. Each pair of horizontal and vertical convolution kernels is individually applied to each input channel. Then, pointwise convolution is carried out, convolving across all input channels simultaneously. By dividing single convolution operation into two parts, axial depthwise convolution enables a massive reduction of parameters. 

\IncMargin{1em}
\begin{algorithm} 
	\KwIn{A 3-dimension tensor $In$ of $l$ channels} 
	\KwOut{A 3-dimension tensor $Out$ of $l$ channels}
	 \BlankLine 
	 
      


    \For{$i\leftarrow 0$ \KwTo $l$}{
        $Hconv\left[ i \right]$ = Conv($In\left[i \right]$, $horizontal = True$)\;
        $Vconv\left[ i \right]$ = Conv($In\left[ i \right]$, $vertical = True$)\;
        $DWconv[i]$ = $Hconv\left[ i \right]$ + $Lconv\left[ i \right]$ + $In[i]$\;
    }
    $Out$ = Conv($DWconv$, $all = True$)
 	 	  \caption{Axial Depthwise Convolution}
 	 	  \label{ADW} 
 	 \end{algorithm}
 \DecMargin{1em}


\subsection{Channel Attention Module}
The channel attention module \cite{ref34} focuses on dynamically adjusting the weights of each channel based on its features. This module offers an effective
solution for channel-specific degradation in underwater images.
By incorporating the channel attention module, our model can learn the significance of different channels and assign higher weights to the most informative ones. 
Thus, we can reduce the inconsistent attenuation by addressing the varying degradation levels among channels. 

\begin{figure}[ht]
    \centering
    \includegraphics[scale=0.35]{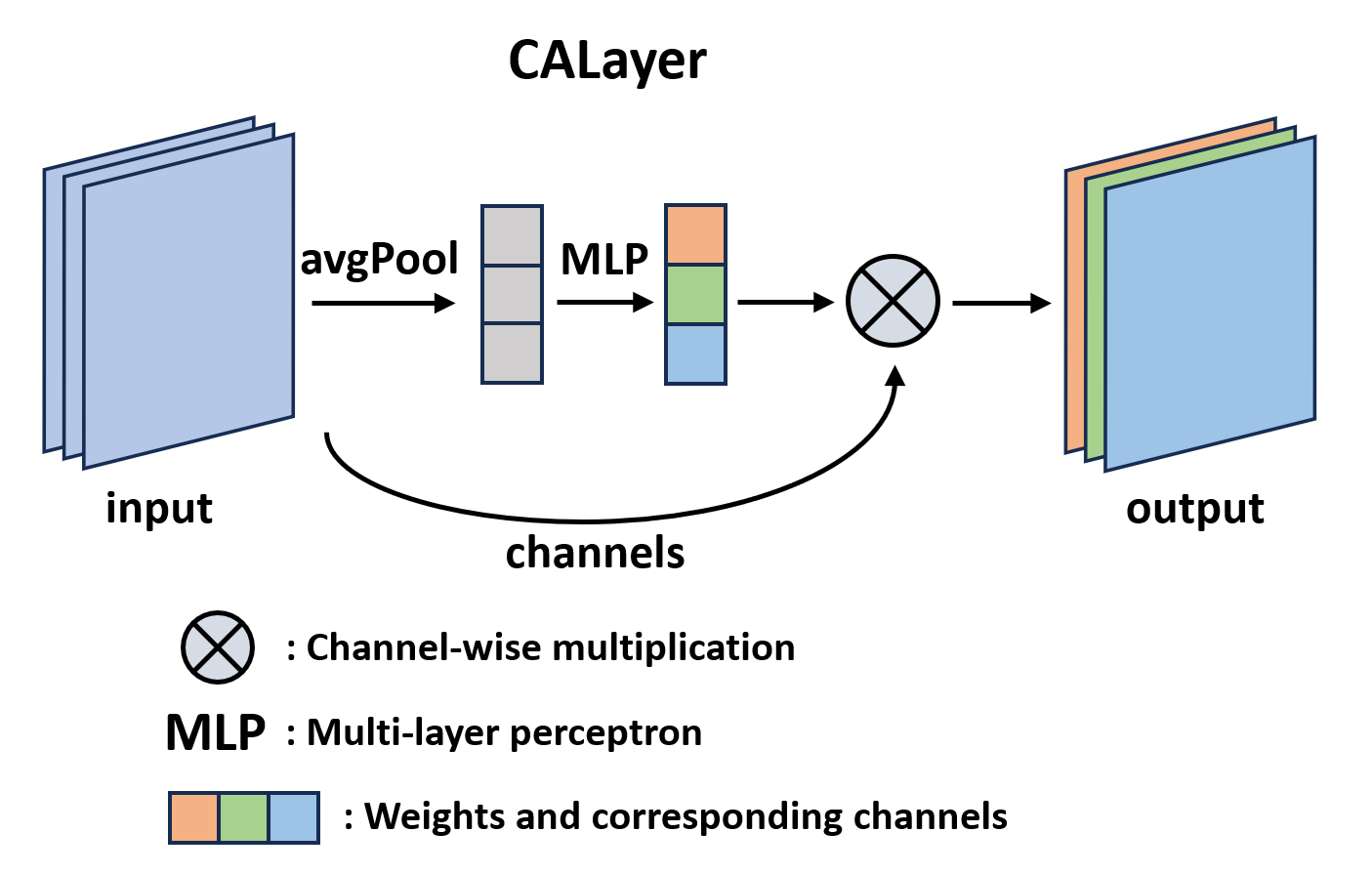}
    \vspace{-18pt}
    \caption{The illustration of CALayer based on channel attention module. The features of input channels are first extracted by average pooling and then adjusted by 
    MLP-like architecture for channel weights. The adaptation of channels is carried out by element-wise multiplication of input channels and final weights.}
    \label{fig-CALayer}
\end{figure}

The channel attention module comprises three main components.
Firstly, global average pooling is performed on each channel, reducing the two-dimensional features to a single scalar weight value for each channel.
Next, through an multi-layer perceptron (MLP) or a similar MLP-like architecture, the excitation process modifies the original weights to generate channel-wise weights.
Finally, element-wise multiplication is applied between each channel and its corresponding weight, resulting in the final output. The process is represented in Fig. \ref{fig-CALayer}. 

\IncMargin{1em}
\begin{algorithm} 
	\KwIn{A 3-dimension tensor $In$ of $l$ channels} 
	\KwOut{A 3-dimension adjusted tensor $Out$ of $l$ channels}
	 \BlankLine 
	 
      


    \For{$i\leftarrow 0$ \KwTo $l$}{
        $w_{0}\left[ i \right]$ = avgPool($In\left[i \right]$)\;
    }
    $w$ = MLP($w_{0}$)\;
    \For{$i\leftarrow 0$ \KwTo $l$}{
        $Out\left[ i \right]$ = $w\left[ i \right] \times In[i]$\;
     }
    
 	 	  \caption{CALayer}
 	 	  \label{CA} 
 	 \end{algorithm}
 \DecMargin{1em}

\subsection{Loss Function Design}
To attain a more comprehensive sense of image information and achieve a more pleasing visual effect, we design a loss function expressed as follows:
\begin{equation}\label{lf}
        L_{total} = l_{RGB}+l_{LAB}+l_{LCH}+l_{SSIM}+l_{VGG}.
\end{equation}
$l_{RGB}$, $l_{LAB}$, $l_{LCH}$ are mean square error in RGB, LAB and LCH color spaces. $l_{SSIM}$ is the SSIM loss. $l_{VGG}$ is the VGG loss. SSIM (Structural Similarity Index) evaluates the similarity of luminance, contrast and structure between output image and the ground truth. VGG loss compares the high-level features of two images after the process of the pretrained VGG network.  

In the case of underwater image enhancement, where datasets are relatively small, it is essential to incorporate comprehensive image information from various perspectives to achieve superior performance. Therefore, our proposed loss function (\ref{lf}) leverages multiple image properties to encompass a broader range of image information, resulting in improved performance.

SSIM is a widely adopted metric for evaluating image quality, where a higher SSIM score typically indicates superior image quality. In line with the experimental findings reported in \cite{ref13}, it was observed that the RGB, LAB, and LCH color spaces yield the highest SSIM scores when applied to underwater images. Therefore, we incorporate these three color spaces and the SSIM loss into our loss function, incorporating multiply image characteristics.

Furthermore, to achieve a more perceptually pleasing enhancement result, we employ the VGG loss. The application of VGG loss ensures more satisfactory visual perceptual effects of enhanced images.

\section{EXPERIMENTS}\label{sec-ex}
In this section, we begin by providing an overview of the experimental settings about the processing hardware and the underwater robotic platform on which 
the real-world experiments are conducted. 
Following this, training specifics for our Lightweight Underwater UNet model are presented. 
Then, training experiments conducted on a dataset compare the enhancement performance and processing speed of 
our model with other state-of-the-art models. Moreover, our model is integrated into the underwater remotely operated vehicle (ROV) to test the visual 
effects of enhanced images and processing speed in real-world underwater tasks.

\subsection{Experimental Settings}
Our experiments are conducted on an RTX 3060 laptop. Besides, the extensive experiment of LU2Net is carried out 
on an i7-10750H processor to test the speed without a graphics processing unit. 

\begin{figure}[thpb]
    \centering
    \includegraphics[scale=0.35]{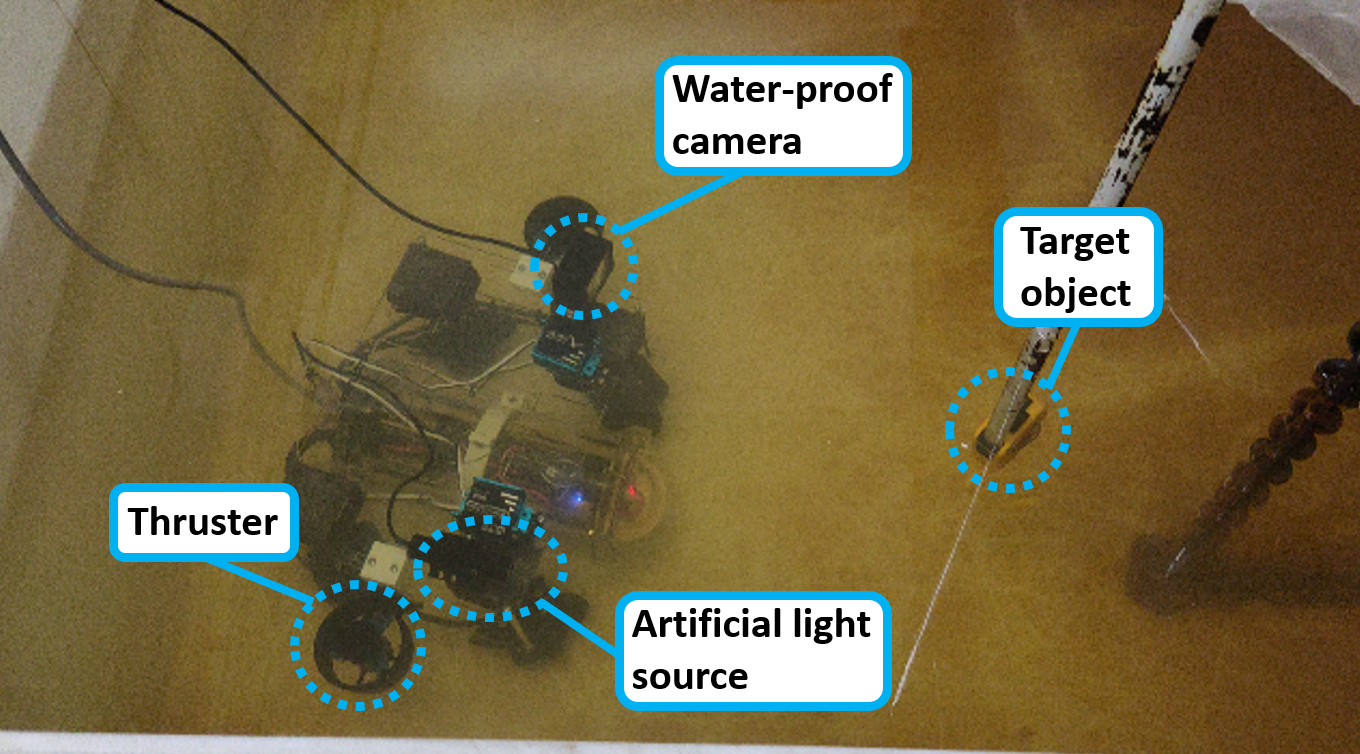}
    \caption{Our underwater ROV and experimental environment}
    \label{fig-env1}
\end{figure}

Fig. \ref{fig-env1} illustrates the structure of the underwater ROV for real-world experiments and the experimental environment. Our underwater ROV is 
equipped with a water-proof camera for visual perception, which can output videos of 640$\times$480  resolution at the speed of 30 frames per second (fps). 
The artificial light source provides visible light in poor lightning conditions. Thrusters attached to the body empower our ROV with motion ability.

\subsection{Training Details}
We use a comprehensive underwater image enhancement dataset \cite{ref13}, LSUI, which consists of 5000 pairs of original input 
images and corresponding ground truth images. This dataset significantly surpasses previous datasets in terms of size, making it an 
ideal choice for our experiments.

For the implementation of our model, we utilize the Python programming language and the PyTorch framework. Throughout the entire 
training process, which spans 150 epochs, we employ the Adam optimization algorithm. The initial learning rate is set to 0.0005, 
and every 40 epochs, the learning rate is reduced by 20 percent.

\subsection{Model Evaluation}
As is customary, we partition the LSUI dataset into a training dataset and a testing dataset, following an 8:2 split. The images 
in the dataset undergo resizing to dimensions of $256\times256\times3$, and normalization is performed to ensure pixel values fall within the range of $[-1,1]$.

To evaluate the enhancement performance and processing efficiency of our proposed method, we compare it with recent state-of-the-art 
UIE methods. Each model undergoes training on the designated training dataset for a total of 150 epochs.

During the comparison, we assess various metrics, including image quality, processing time, and parameter count, to provide a 
comprehensive evaluation of different methods. 

\subsubsection{Enhancement Performance Evaluation}

\begin{figure*}
        \centering
        \subfigure[Real-world images]{
        \begin{minipage}[b]{0.18\linewidth}
        \includegraphics[width=1\linewidth]{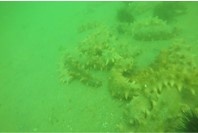}\vspace{2pt}
        \includegraphics[width=1\linewidth]{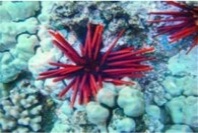}\vspace{2pt}
        \includegraphics[width=1\linewidth]{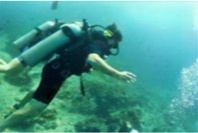}
        \end{minipage}}
        \subfigure[MLLE]{
        \begin{minipage}[b]{0.18\linewidth}
            \includegraphics[width=1\linewidth]{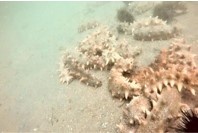}\vspace{2pt}
            \includegraphics[width=1\linewidth]{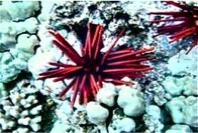}\vspace{2pt}
            \includegraphics[width=1\linewidth]{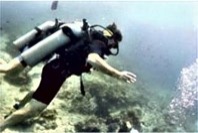}
        \end{minipage}}
        \subfigure[U-Trans]{
        \begin{minipage}[b]{0.18\linewidth}
            \includegraphics[width=1\linewidth]{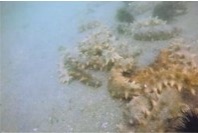}\vspace{2pt}
        \includegraphics[width=1\linewidth]{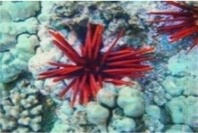}\vspace{2pt}
        \includegraphics[width=1\linewidth]{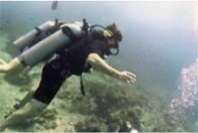}
        \end{minipage}}
        \subfigure[Ours]{
        \begin{minipage}[b]{0.18\linewidth}
            \includegraphics[width=1\linewidth]{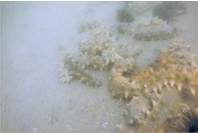}\vspace{2pt}
        \includegraphics[width=1\linewidth]{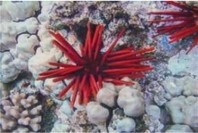}\vspace{2pt}
        \includegraphics[width=1\linewidth]{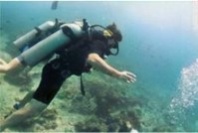}
        \end{minipage}}
        \subfigure[Ground truth]{
        \begin{minipage}[b]{0.18\linewidth}
            \includegraphics[width=1\linewidth]{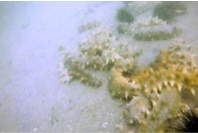}\vspace{2pt}
        \includegraphics[width=1\linewidth]{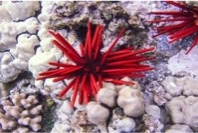}\vspace{2pt}
        \includegraphics[width=1\linewidth]{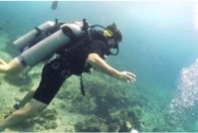}
        \end{minipage}}
        \caption{Illustration of enhancement results by different models}
        \label{fig-dataset_expriment}
        \end{figure*}

        MLLE serves as an example of a non-deep-learning approach for underwater image enhancement, 
        while U-Trans stands out as a recently proposed deep learning method that achieves superior results. Additionally, UWCNN and WaterNet are notable examples of previous state-of-the-art UIE models.

        MLLE inherits the limitations associated with traditional UIE methods, including color distortion and an inability to handle diverse underwater color and lighting conditions. In contrast, both U-Trans 
        and our proposed model generate visually appealing enhanced images. These enhanced images enable improved identification of underwater objects and the surrounding environment, thereby facilitating 
        higher-level tasks such as object tracking, object detection, and obstacle avoidance.

        From the comparative analysis presented in Table. \ref{tab-performance} and the illustration of enhanced underwater images in Fig. \ref{fig-dataset_expriment}, it is evident that our model attains a state-of-the-art level of enhancement performance 
        when compared to the selected methods. The evaluation metrics used include PSNR, SSIM and subjective visual quality assessment. Our model 
        outperforms the other methods in terms of these metrics, indicating its superiority in enhancing underwater images. This achievement can be attributed to the effective utilization of advanced deep 
        learning techniques and the incorporation of domain-specific knowledge in our model's architecture.

\begin{table}[ht]\label{tab-performance}
        \caption{Comparisons of enhancement performance}
        \label{1}
        \vspace{-8pt}
        \begin{center}
        \begin{tabular}{|c|c|c|c|}
        \hline
        Method & PSNR & SSIM \\
        \hline
        MLLE & 18.133 & 0.730   \\
        \hline
        WaterNet & 19.517 & 0.813   \\
        \hline
        UWCNN & 22.621 & 0.825   \\
        \hline
        U-Trans & 25.032 & 0.843  \\
        \hline
        Ours & \textbf{25.549} & \textbf{0.868}  \\
        \hline
        \end{tabular}
        \end{center}
        \end{table}

\subsubsection{Model Speed and Parameters Comparison}

The experimental results demonstrate that our model achieves a remarkable processing speed. Utilizing a common RTX 3060 laptop, 
our model is capable of outputting videos at 100 fps, thereby meeting 
the real-time demand of underwater tasks. Additionally, without a GPU, our model can still generate enhanced images at 
the speed of 12 fps on an i7-10750H processor.

Moreover, the experiment results exhibits the lowest demand for computational resources of our model, as indicated by 
the lowest FLOPs (Floating Point Operations) requirement. Furthermore, with a relatively small number of parameters, our model 
does not necessitate a large amount of memory. Consequently, our model is lightweight enough to be deployed on underwater robotic 
platforms while achieving impressive enhancement results. As a low-level underwater image enhancement solution, our model preserves 
ample computational resources for other tasks and coexists harmoniously with high-level vision models. 

\begin{table}[th]\label{tab-speed}
        \caption{Comparisons of processing speed and parameters}
        \label{2}
             \vspace{-8pt}
        \begin{center}
        \begin{tabular}{|c|c|c|c|}
        \hline
        Method & FLOPs & Parameters& Time/Frame\\
        \hline
        MLLE & / & / & 0.163s  \\
        \hline
        WaterNet & 143.3G & 1.0M &0.475s \\
        \hline
        UWCNN & \textbf{5.2G} & \textbf{40.0K} & \textbf{0.005s}\\
        \hline
        U-Trans & 60.2G & 22.8M & 0.08s\\
        \hline
        Ours & \textbf{2.8G} & \textbf{176K} & \textbf{0.01s}\\
        \hline
        \end{tabular}
        \end{center}
        \end{table}

\subsection{Real-world Test}

To test the performance of our model on real-world robots, we conduct a real-time underwater image enhancement experiment on our underwater 
vision-driven ROV. Due to the limitation of processing speed of different models, we only compare the performance of UWCNN 
and our model that can handle real-time enhancement tasks. In absence of ground truth, PSNR and SSIM are not suitable for 
the assessment of real-world underwater image enhancement. Instead, we choose UCIQE \cite{ref37} as the alternative metric, which combines chroma, saturation,
and contrast for image quantification. Commonly, higher UCIQE value indicates better image quality. 

\begin{figure}
    \centering
    \subfigure[Real-world]{
    \begin{minipage}[b]{0.31\linewidth}
    \includegraphics[width=1\linewidth]{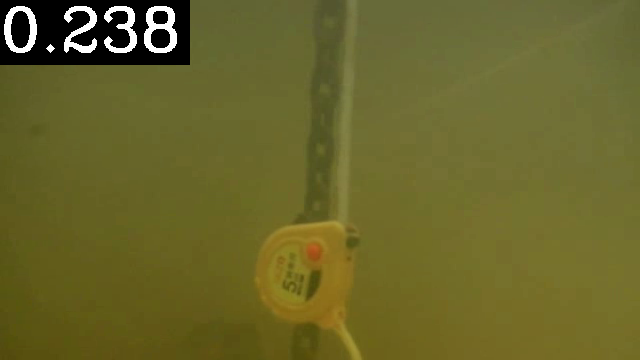}\vspace{1pt}
    \includegraphics[width=1\linewidth]{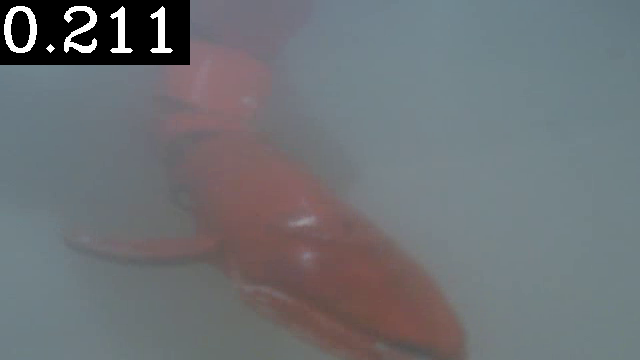}\vspace{1pt}
    \includegraphics[width=1\linewidth]{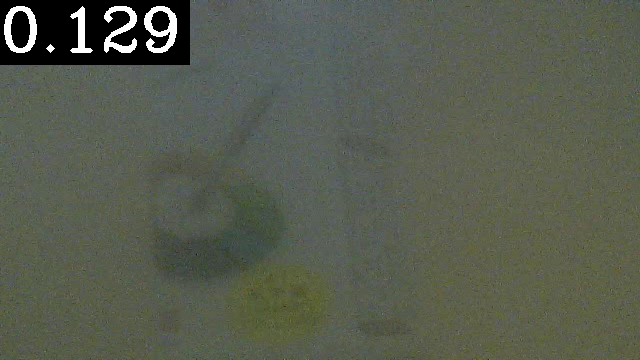}
    \end{minipage}}
    \subfigure[UWCNN]{
    \begin{minipage}[b]{0.31\linewidth}
    \includegraphics[width=1\linewidth]{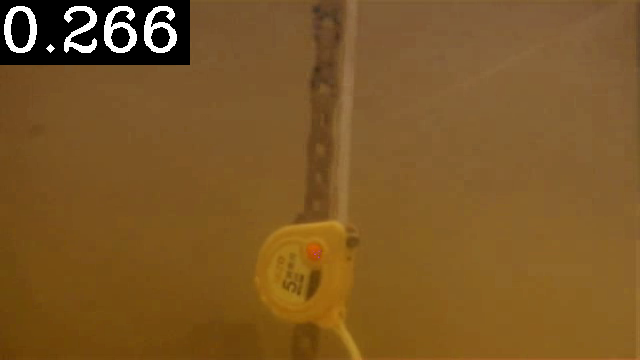}\vspace{1pt}
    \includegraphics[width=1\linewidth]{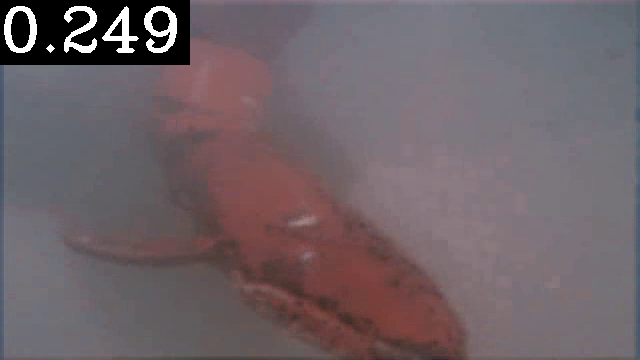}\vspace{1pt}
    \includegraphics[width=1\linewidth]{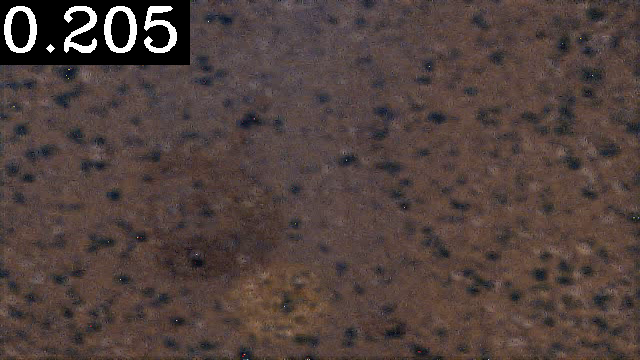}
    \end{minipage}}
    \subfigure[Ours]{
    \begin{minipage}[b]{0.31\linewidth}
    \includegraphics[width=1\linewidth]{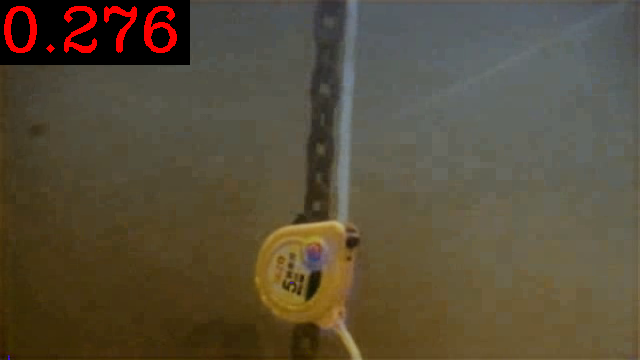}\vspace{1pt}
    \includegraphics[width=1\linewidth]{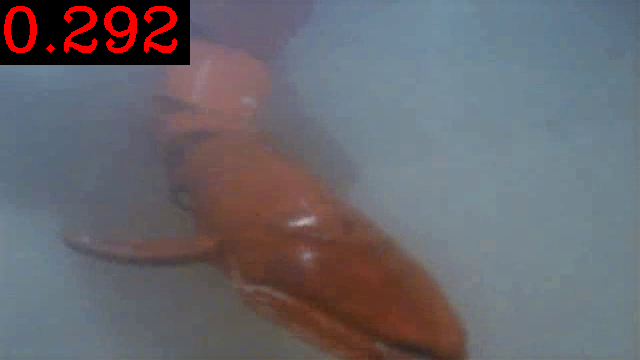}\vspace{1pt}
    \includegraphics[width=1\linewidth]{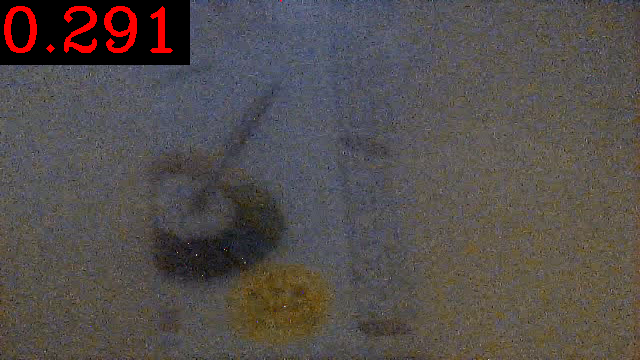}
    \end{minipage}}
    \caption{Real-world enhancement results by real-time models. The UCIQE value is indicated in the top left corner of each image.}
    \label{fig-real_world_experiment}
    \end{figure}

Real-world experiments shown in Fig. \ref{fig-real_world_experiment} prove the ability of LU2Net to provide high-quality enhanced underwater videos. The output video of 30 fps by the 
camera is immediately enhanced without noticeable delay, showcasing the potential of our model for real-time marine vision tasks. 

From the illustrated experiment results, our model achieves a significant visual effect and outputs balanced and 
detailed images from a subjective aspect. Moreover, higher UCIQE values further prove the significant performance of LU2Net. With our model, 
human operators can clearly identify underwater objects and environments 
through well-enhanced videos and make proper decisions for better robotic safety and performance.

\section{CONCLUSIONS}

In this paper, we proposed a lightweight underwater image enhancement model that demonstrated superior capabilities in effectively removing image distortion and 
providing real-time enhancement.
Our model combined axial depthwise convolution, the channel attention module, and the U-Shape net structure, resulting in the lightweight net structure and significantly fast processing speed. The experiments conducted on the dataset and 
real-world robots confirmed the model's low computational requirements and efficient resource utilization, which enabled the proposed model to handle real-time image and video enhancement. 
This advantage enhanced our model's suitability for integration into underwater robots for real-time vision tasks.



\bibliographystyle{IEEEtran}
\bibliography{IEEEabrv,ref}

\end{document}